\documentclass[11pt,a4paper]{article}
\usepackage[hyperref]{naaclhlt2019}
\usepackage{times}
\usepackage{latexsym}
\usepackage{graphicx}

\usepackage{amsmath,amssymb,mathtools}

\usepackage{todonotes}

\newcommand{\specialcell}[2][c]{%
  \begin{tabular}[#1]{@{}c@{}}#2\end{tabular}}
\usepackage{tabularx}
\usepackage{boldline}

\usepackage{url}

\aclfinalcopy 
\def\aclpaperid{7} 

\title{A Framework for Decoding Event-Related Potentials from Text} 

\author{Shaorong Yan \\
  Department of Brain and Cognitive Sciences\\ University of Rochester\\  Rochester, NY 14627, USA  \\
  {\tt syan13@ur.rochester.edu} \\\And
  Aaron Steven White \\
  Deparment of Linguistics\\ University of Rochester\\ Rochester, NY 14627, USA \\
  {\tt aaron.white@rochester.edu} \\}
\date{}

\begin{document}

\maketitle
\begin{abstract}
We propose a novel framework for modeling event-related potentials (ERPs) collected during reading that couples pre-trained convolutional decoders with a language model. Using this framework, we compare the abilities of a variety of existing and novel sentence processing models to reconstruct ERPs.  We find that modern contextual word embeddings underperform surprisal-based models but that, combined, the two outperform either on its own. 
\end{abstract}

\section{Introduction}

Understanding the mechanisms by which comprehenders incrementally process linguistic input in real time has been a key endeavor of cognitive scientists and psycholinguists. Due to its fine time resolution, event-related potentials (ERPs) are an effective tool in probing the rapid, online cognitive processes underlying language comprehension. Traditionally, ERP research has focused on how the properties of the language input affect different ERP components \citep[see][for reviews]{van2012,kuperberg2016}.\footnote{Examples of such components include the N1/P2 \citep{sereno1998,dambacher2006}; N250 \citep{grainger2006}; N400 \citep{kutas1980, hagoort2004,lau2008}; and P600 \citep{osterhout1992, kuperberg2003, kim2005}}

While this approach has been fruitful, researchers have also long been aware of the potential drawbacks to this \textit{component-centric} approach: a predictor's effects can be too transient to detect when averaging ERP amplitudes over a wide time window---as is typical in component-based approaches \citep[see][for discussion]{hauk2006}. Different predictors can affect ERP in the same time window as an established component but have slightly different temporal \citep{frank2017} or spatial \citep{delong2005} profiles. This means that the definition of a component strongly affects interpretation.

\begin{figure}[tb!]
\centering
\includegraphics[width=\columnwidth]{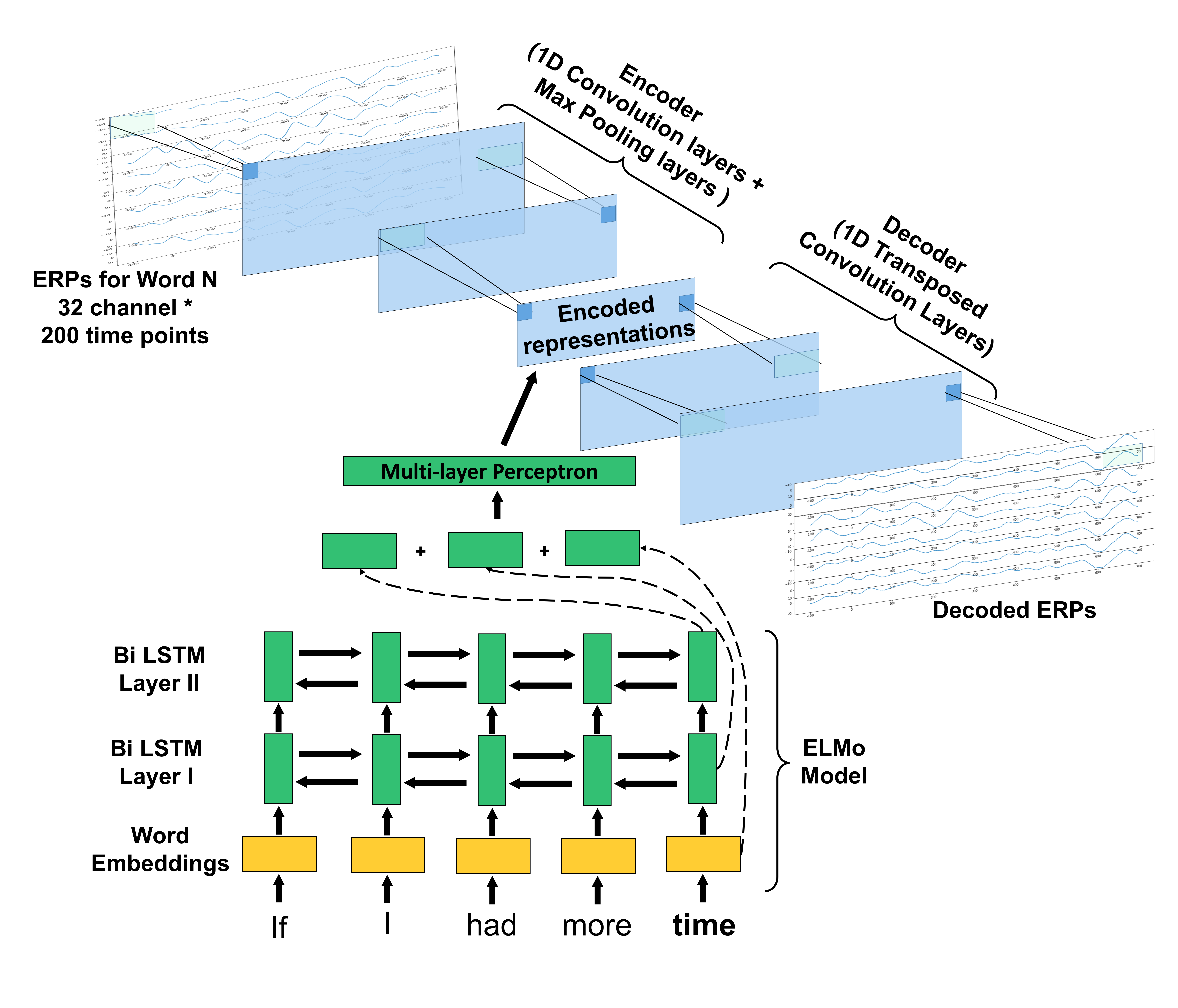}
\vspace{-10mm}
\caption{An instance of our framework using a bidirectional language model as the text encoder.}
\label{fig:Model_Structure}
\vspace{-6mm}
\end{figure}

There are two typical approaches to resolving these issues. The first is to plot the data and use visual inspection to select an analysis plan, introducing uncontrollable researcher degrees of freedom \citep{gelman2014}. Another approach is to run separate models for each time point (or even each electrode) to look for the emergence of an effect. This necessitates complex statistical tests to monitor for inflated Type I error \citep[see, e.g.,][for discussion]{blair1993, laszlo2014} and to control for autocorrelation across time points \citep{smith2015regression,smith2015}.

We explore an alternative approach to the analysis of ERP data in language studies that substantially reduces such researcher degrees of freedom: directly decoding the raw electroencephalography (EEG) measurements by which ERPs are collected. Inspired by multimodal tasks like image captioning \citep[see][for a review]{hossain_comprehensive_2019} and visual question answering \citep{antol2015}, we propose to model EEG using standard convolutional neural networks (CNNs) pre-trained under an autoencoding objective. The decoder CNN can then be decoupled from its encoder and recoupled with any language processing model, thus enabling explicit quantitative comparison of such models. We demonstrate the efficacy of this framework by using it to compare existing sentence processing models based on surprisal and/or static word embeddings with novel models based on contextual word embeddings. We find that surprisal-based models actually outperform contextual word embeddings on their own, but when combined, the two outperform either model alone. 

\section{Models}

All of the models we present have two components: (i) a pre-trained CNN for decoding raw EEG measurements time-locked to each word in a sentence; and (ii) a language model from which features can be extracted for each word---e.g. the surprisal of that word given previous words or its contextual word embedding. An example model structure using ELMo embeddings \citep{peters2018} is illustrated in Figure \ref{fig:Model_Structure}.

\paragraph{Convolutional decoder} 

For all models, we use a convolutional decoder pre-trained as a component of an autoencoder. To reduce researcher degrees of freedom, the decoder architecture is selected from a set of possible architectures by cross-validation of the containing autoencoder.

The autoencoder consists of two parts: (a) a convolutional encoder that finds a way to best compress the ERP signals; and (b) a convolutional decoder with a homomorphic architecture that reconstructs the ERP data from the compressed representation. ERPs were organized into a 2D matrix (channel $\times$ time points). For the encoder, we pass the ERPs through multiple interleaved 1D convolutional and max pooling layers with receptive fields along the time dimension, shrinking the number of latent channels at each step. Correspondingly, for the decoder part, we use a homomorphic series of 1D transposed convolutional layers to reconstruct the ERP data. 

At train time, the decoder weights are frozen, and the encoder is replaced by one of the language models described below. This entails fitting an \textit{interface mapping}---a linear transformation for each channel produced by the encoder---from the features extracted from the language model into the representation space output by the encoder.

\paragraph{Language models}

We consider a variety of features that can be extracted from a language model.

\subparagraph{Surprisal}

We use the lexical surprisal $-\log p(w_i\;|\;w_1, \ldots, w_{i-1})$ obtained from a RNN trained by \citet{frank2015}.

\subparagraph{Semantic distance}

Following \citet{frank2017}, we point-wise average the GloVe embedding \citep{pennington2014} of each word prior to a particular word to obtain a context embedding and then calculate the cosine distance between the context embedding and the word embedding for that word. We use the GloVe embeddings trained on Wikipedia 2014 and Gigaword 5 (6B tokens, 400K vocabulary size).

\subparagraph{Static word embeddings} 

We also consider the GloVe embedding dimensions as features. We do not tune the GloVe embeddings using an additional recurrent neural network (RNN), instead just passing the them through a multi-layer perceptron with one hidden layer of tanh nonlinearities. The idea here is that the GloVe-only model tells us how much the distributional properties of a word, outside of the current context, contribute to ERPs.

\subparagraph{Contextual word embeddings} 

We consider contextual word embeddings generated from ELMo \citep{peters2018} using the \texttt{allennlp} package \cite{gardner2017}. ELMo produces contextual word embeddings using a combination of character-level CNNs and bidirectional RNNs trained against a language modeling objective, and thus it is a useful contrast to GloVe, since it captures not only a word's distributional properties, but how they interact with the current context. 

We take all three layers of the hidden layer output in the ELMo model and concatenate them. To ensure a fair comparison with the surprisal- and GloVe-based models, we use the same tuning procedure employed for the static word embeddings. Further, because sentences are presented incrementally in ERP experiments and because ELMo is bidirectional and thus later words in the sentence will affect the word embeddings of previous words, we do not obtain an embedding for a particular word on the basis of the entire sentence, instead using only the portion of the sentence up to and including that word to obtain its embedding.

\subparagraph{Combined models}

We also consider models that combine either static or contextual word embedding features with frequency, surprisal, and semantic distance. The latter features were concatenated onto the tuned word embeddings before being passed to the interface mapping.

\section{Experiments}
\label{sec:experiments}

We use the EEG recordings collected and modeled by \citet{frank2017}. In their study, 24 subjects read sentences drawn from natural text. Sentences were presented word-by-word using a rapid serial visual presentation paradigm. 
We use the ERPs of each word epoched from -100 to 700ms and time-locked to word onset from all the 32 recorded scalp channels. After artifact rejection (provided by \citeauthor{frank2017} with the data), this dataset contains 41,009 training instances.

\begin{figure}[tb!]
\centering
\includegraphics[width=0.9\columnwidth]{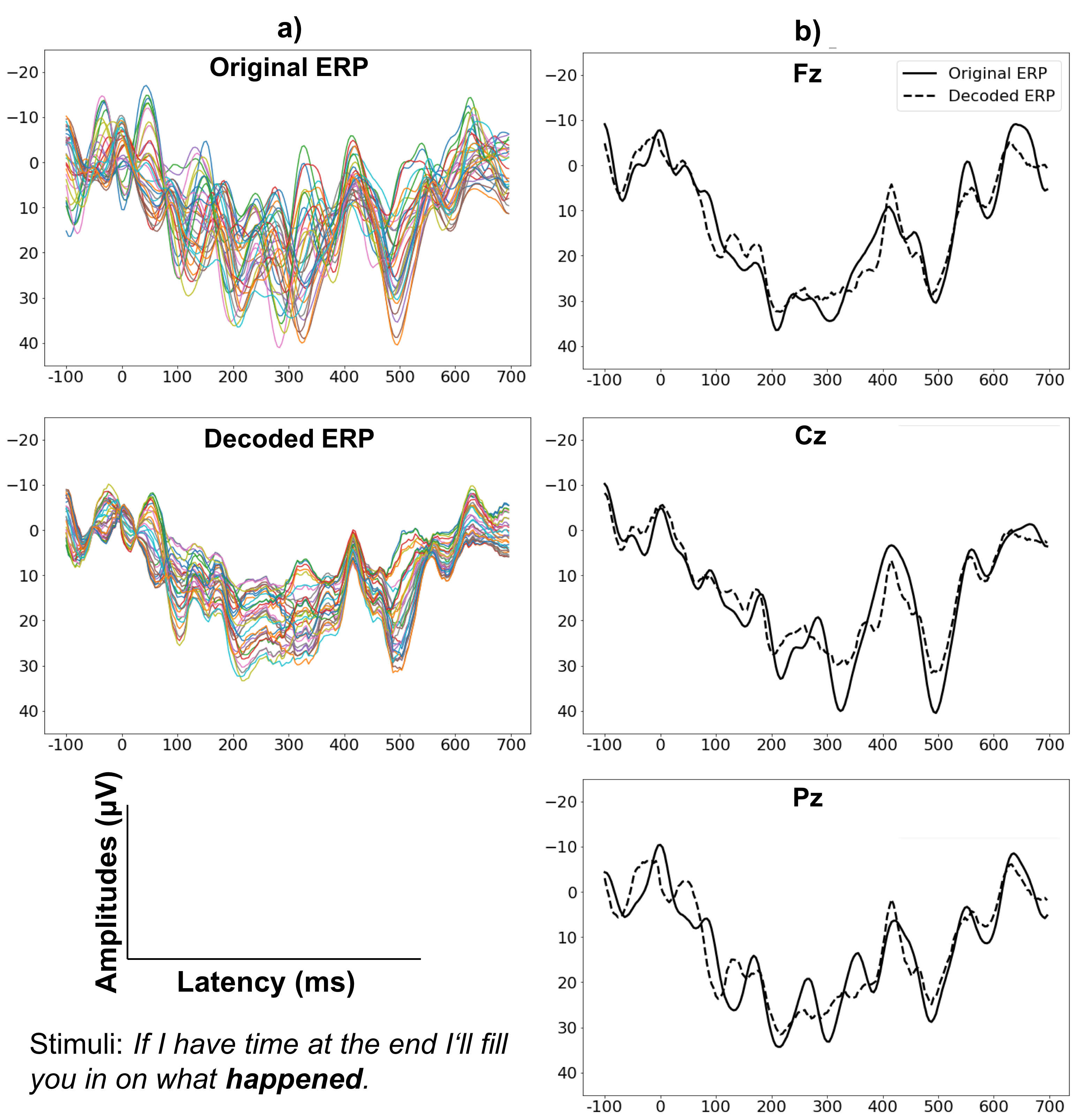}
\vspace{-3mm}
\caption{Original ERPs and ERPs decoded from the trained autoencoder of an example trial. a) ERPs from all 32 channels (denoted by color). b) Original (solid) and decoded ERPs (dashed) for example electrodes.}
\label{fig:autoencoder}
\vspace{-6mm}
\end{figure}

\paragraph{Pre-training}

To select which decoder to use, we compare the performance of two CNN architectures motivated by well-known properties of EEG. The first architecture has $5$ latent channels and $9$ time steps. Given the sampling rate and size of the input ($250$Hz, $200$ time steps), this roughly corresponds to filtering the EEG data with alpha band frequency ($\sim10$Hz). The other has $10$ latent channels and $20$ time steps, thus lying within the range of beta band activity ($\sim25$Hz). In addition to these two architectures, we also examine whether including subject- and electrode-specific random intercepts improves model performance. 

We conduct a 5-fold cross-validation for each architecture to find the one that has the best performance in reconstructing ERP data. As shown in Table \ref{tb:autoencoder_selection}, the beta models perform better overall than alpha models, since they likely capture both alpha and beta band activities. Adding subject-specific intercept, on the other hand, did not greatly improve the model performance. 

\begin{table}[h]
\centering
\begin{tabular}{ccc} 
\hlineB{4}
 Model &  \specialcell{No Intercept} & \specialcell{Intercept}\\
\hlineB{4}
alpha  & $49.9\;(0.532)$	&	$49.7\;(0.533)$\\
beta   & $33.5\;(0.686)$	&	$32.7\;(0.692)$\\
\hlineB{4}
\end{tabular}
\vspace{-2mm}
\caption{Mean MSE and $R^2$ (in parentheses).}
\label{tb:autoencoder_selection}
\vspace{-4mm}
\end{table}

\noindent Figure \ref{fig:autoencoder} shows the reconstructed ERPs of the beta model on one trial. The autoencoder can reconstruct the ERP signal very well. The selected channels are illustrative of the reconstruction accuracy across all channels. We thus selected the beta model without subject-specific intercept as the decoder for our consequent models.

\paragraph{Training}

The interface mapping and (where applicable) word embedding tuner are trained under an MSE loss using mini-batch gradient descent (batch size = 128) with the Adam optimizer (\texttt{learning rate=0.001} and default settings for \texttt{beta1}, \texttt{beta2}, and \texttt{epsilon}) implemented in \texttt{pytorch} \cite{paszke2017}. Each model is trained for 200 epochs. Since we need at least one preceding word to compute contextual word embeddings, we do not include the first word of the sentence. This left ERPs for 1,618 word tokens per subject (638 word types). After excluding trials containing artifacts, a total of 37,112 training instances remain.

\paragraph{Development}

To avoid overfitting, we use early stopping and report the models with the best performance on the development set. We did a parameter search over three different weight decays: \texttt{1e-5}, \texttt{1e-3}, \texttt{1e-1}. For each model, we chose the weight decay that produced the best mean performance on held-out data in a 5-fold cross-validation. 

\paragraph{Baselines}

As a baseline we train an intercept-only model that passes a constant input (optimized to best predict the data) to the decoder. In addition, we fit a baseline model that only has word frequency as a feature. Frequency is also included as an additional feature in all models.

\paragraph{Metrics}

To account for the fact that our model performance is bounded by the performance of the autoencoder, we report a modified form of $R^2$ to evaluate the overall model performance.

\vspace{-6mm}
\[R^2_\text{mod} = 1 - \frac{\text{MSE}_\text{model} - \text{MSE}_\text{autoencoder}}{\text{MSE}_\text{intercept} - \text{MSE}_\text{autoencoder}}\] 
\vspace{-4mm}

\begin{table}[tb!]
\centering
\vspace{-0mm}
\begin{tabular}{ccc} 
\hlineB{4}
 Model &  $R^2_\text{mod}$ & $95$\% CI\\
\hlineB{4}
Frequency                    & $19.5$ & $[18.5, 20.7]$\\
F + Surp            & $37.4$ & $[36.5, 38.3]$ \\
F + SemDis           & $36.1$ & $[32.3, 38.4]$ \\
F + GLoVE                & $35.0$ & $[31.8, 38.2]$ \\
F + ELMo                & $35.2$ & $[34.3, 36.2]$\\
F + S + SD               & $46.6$ & $[43.5, 49.7]$ \\
F + S + SD + GloVe      & $47.1$ & $[43.2, 49.4]$ \\
F + S + SD + ELMo       & $49.5$ & $[48.9, 50.1]$ \\
\hlineB{4}
\end{tabular}
\caption{Proportion variance explained by each model ($\times$100) and confidence interval across folds computed by a nonparametric bootstrap. F = frequency, S(urp) = surprisal, S(em)D(is) = semantic distance. }
\label{tb:model_performance}
\vspace{-6mm}
\end{table}

\section{Results}

Table \ref{tb:model_performance} shows the $R^2_\text{mod}$ metric for each model. We see that both surprisal and semantic distance outperform both types of word embedding features, all of which outperform frequency alone. When combined, surprisal and semantic distance outperform either alone, and further gains can be made with the addition of either static (GloVe) or contextual (ELMo) embedding features. The addition of contextual embedding features increases performance more than the addition of static word embedding features, such that there is some benefit to capturing context over and above that provided by surprisal and semantic distance. 

\paragraph{Time course analysis}

\begin{figure}[tb!]
\centering
\includegraphics[width=\columnwidth]{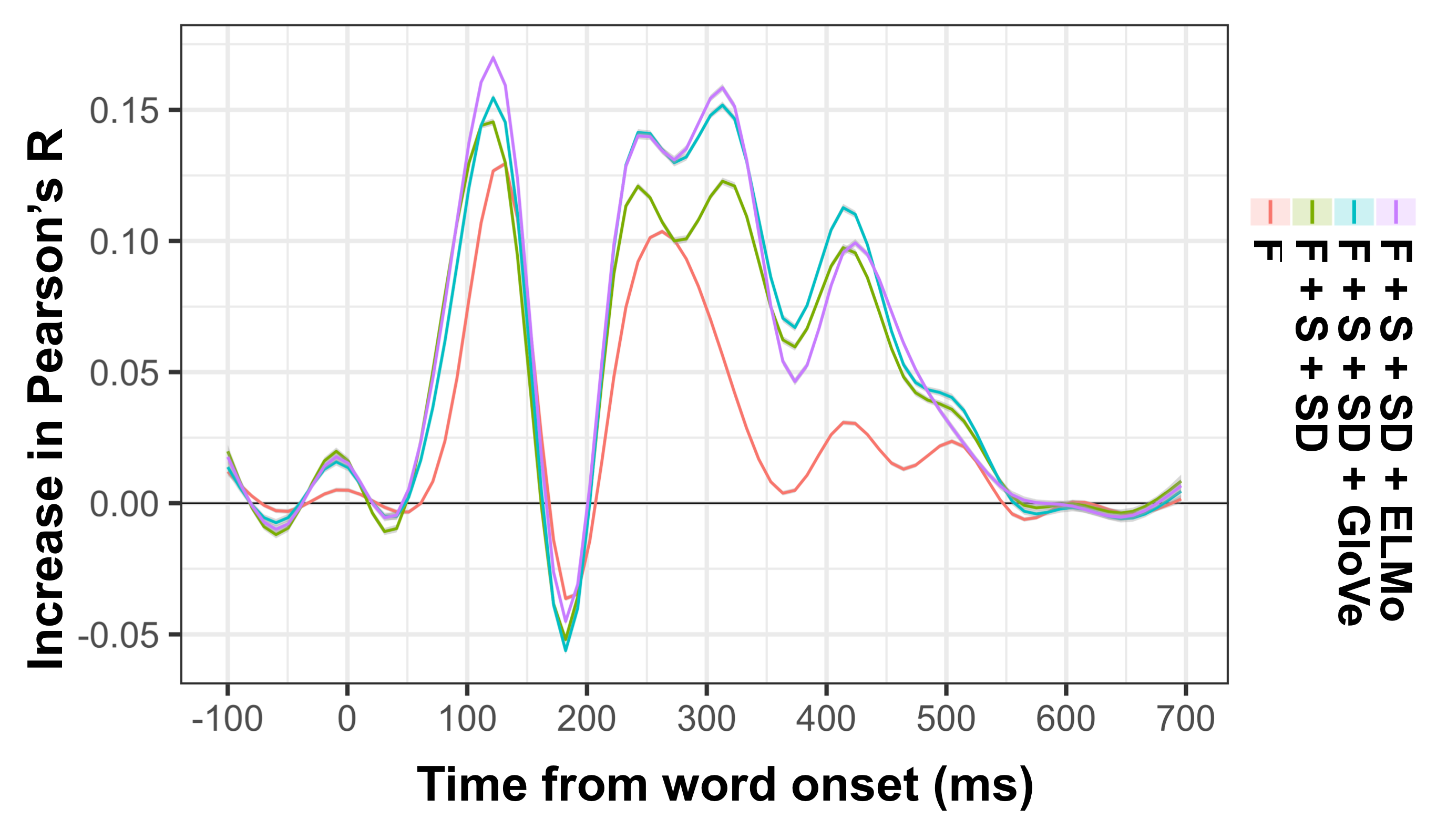}
\vspace{-8mm}
\caption{Increase in Pearson's R between predicted and actual ERPs. Lines show GAM smooth over time.}
\label{fig:PearsonR}
\vspace{-6mm}
\end{figure}

To understand where in time each predictor improved model performance, we examine the increase in correlation over the intercept model at each time point (Figure \ref{fig:PearsonR}). There are roughly three regions where the language models outperform the intercept model. The first is right after 100ms post word onset: corresponding to the N1 component, which is typically considered to reflect perceptual processing; the second is between 200 and 350ms: corresponding to the N250 component, which correlates with lexical access \cite{grainger2006,laszlo2014}; and the third is between 300ms and 500ms: corresponding to the N400, which is typically associated with semantic processing.\ 

Consistent with previous findings \citep{hauk2006,laszlo2014,yan2019}, adding frequency into the model improved model performance in all three time windows. Also consistent with the literature, adding surprisal and semantic distance improved model performance in the N400 time window \citep{frank2017, yan2019}.  

Models with word embeddings do not differ much from the models containing only frequency, surprisal, and semantic distance, with the biggest difference around 300ms post word onset. This might indicate an effect in the early N400 time window. This could also indicate that processes commonly associated with the N250 may be better captured by the models containing word embeddings. If so, it is less expected and potentially interesting, since most of our models have no access to perceptual properties of the input---with the possible exception of ELMo, whose charCNN may capture orthographic regularities. These effects could reflect our models' ability to capture top-down lexical processing \cite[see, e.g.,][]{penolazzi2007, yan2019} or possibly systematic correlations between higher-level features and perceptual features.\

\paragraph{Part-of-speech analysis}

Prior work on ERP during reading distinguishes function word---such as determiners, conjunctions, pronouns, prepositions, numerals, particles---from content words---such as (proper) nouns, verbs, adjectives, adverbs \citep{nobre1994,frank2015}. As such, we also examine whether each model's performance differs for content words and function words. We calculate the Pearson's correlation between the predicted and actual ERPs for each word of each model and used linear mixed-effects model to examine the influence on model fit with the inclusion of different information. If a model included a specific type of information, the corresponding predictor is coded as 1, otherwise it was coded as -1. For example, the surprisal model was trained with surprisal but not semantic distance, so the surprisal predictor is 1 for this model and the semantic distance predictor is -1. We further included the interaction between models and word types (function=-1, content=1).

Table \ref{tb:model_pos} shows the resulting coefficients. Overall, models display better performance for content words than for function words ($\hat{\beta} = 0.003 \text{, t} = 2.01 \text{, \textit{p}} < 0.05$), consistent with previous findings \citep{frank2015}. Including each type of information also significantly increased model fit ($\text{ts} > 10.1 \text{, \textit{p}} < 0.01$). There was a significant interaction between frequency and word type ($\hat{\beta} = -0.001 \text{, t} = -2.43 \text{, \textit{p}} < 0.02$): including frequency increased model performance for function words more than for content words. There was also a marginally significant interaction between ELMo and word type ($\hat{\beta} = 0.0007 \text{, t} = -1.85 \text{, \textit{p}} < 0.064$), suggesting that including ELMo embeddings increased model performance for content words more than for function words. 

\begin{table}[t]
\centering
\begin{tabular}{crrl} 
\hlineB{4}
\multicolumn{1}{c}{Predictor} & \multicolumn{1}{c}{$\hat{\beta}$}  & \multicolumn{1}{c}{t}  & \\   
\hlineB{4}
Intercept                   & $-0.0013$    &  $-0.225$ &  \\
Word Type (Content)        & $0.0030$    &  $2.01$ & $^{*}$ \\
Frequency                   & $0.0110$     &  $21.2$ & $^{**}$\\ 
Surprisal                   & $0.0050$     &  $13.0$ & $^{**}$ \\
Semantic Distance           & $0.0040$     &  $11.60$ & $^{**}$ \\
GloVe Embeddings            & $0.0040$     &  $10.3$ & $^{**}$ \\ 
ELMo Embeddings             & $0.0040$     &  $10.1$ & $^{**}$ \\
Freq : Word Type            & $-0.0010$     &  $-2.43$ & $^{*}$  \\
Surp : Word Type            & $0.0001$   &  $0.24$ &    \\
SemDis : Word Type          & $-0.0003$     &  $-0.70$ & \\
GloVe : Word Type           & $0.0002$   &  $0.55$ &  \\
ELMo : Word Type            & $0.0007$    &  $-1.85$ & $^{+}$ \\ 
\hlineB{4}
\end{tabular}
\caption{Model estimates and t statistics from mixed-effects model. $^{**}: p < 0.01$; $^{*}: p < 0.05$; $^{+}: p < 0.1$}
\label{tb:model_pos}
\vspace{-6mm}
\end{table}

We also examine the interaction between each type of information and each part-of-speech. Overall, the models had worse performance for particles ($\hat{\beta} = -0.017 \text{, t} = -3.37 \text{, \textit{p}} < 0.01$), nouns ($\hat{\beta} = -0.007 \text{, t} = -1.95 \text{, \textit{p}} < 0.051$) and pronouns ($\hat{\beta} = -0.012 \text{, t} = -1.76 \text{, \textit{p}} < 0.08$). Including each type of information increased overall model fit ($\text{ts} > 6.05 \text{, \textit{p}} < 0.01$). While including frequency increased overall model fit, it increased the model fit for verbs less ($\hat{\beta} = -0.003 \text{, t} = -2.04 \text{, \textit{p}} < 0.05$). No other effects reached significance. 

\section{Related Work}

Traditionally, ERP studies of language processing use coarse-grained predictors like cloze rates,
which often lack the precision to differentiate different neural computational models \citep[for discussion, see][]{yan2017, rabovsky2018}. To overcome such limitations, a main line of attack has been to extract measures from probabilistic language models and evaluate them against ERP amplitudes \citep{frank2015,brouwer2017, rabovsky2018, delaney2019, fitz2018, szewczyk2018, biemann2015}.

While prior studies have also predicted ERPs from language model-based features \citep{broderick2018,frank2017,hale2018}, they fit to aspects of the EEG signals that are unlikely to be related to language processing. Our approach threads the needle by first finding abstract structure in the ERPs with a CNN, then using that knowledge in predicting that structure from linguistic features. We are not the first to use CNNs to model EEG/ERPs \citep{lawhern2016,schirrmeister2017,seeliger2018, acharya2018, moon2018}, but to our knowledge, no other work has yet used CNNs for modeling ERPs during reading.

\section{Conclusion}

We proposed a novel framework for modeling ERPs collected during reading. Using this framework, we compared the abilities of a variety of existing and novel sentence processing models to reconstruct ERPs, finding that modern contextual word embeddings underperform surprisal-based models but that, combined, the two outperform either on its own. 

ERP data provides a rich testbed not only for comparing models of language processing, but potentially also for probing and improving the representations constructed by natural language processing (NLP) systems. We provided one example of how such probing might be carried out by analyzing the differences among models as a function of processing time, but this analysis only scratches the surface of what is possible using our framework, especially for understanding the more complex neural models used in NLP.

\section*{Acknowledgments}
We are grateful to Dr. Stefan Frank for sharing the EEG data, sentence materials, and language model predictors. We would also like to thank three anonymous CMCL reviewers as well as the the FACTS.lab and HLP lab at UR for providing valuable feedback on the draft.

\bibliography{CMCL2019}
\bibliographystyle{acl_natbib}

\end{document}